\documentclass{article} 
\usepackage{iclr2021_conference,times}

\usepackage{amsmath,amsfonts,bm}

\def\eqref#1{equation~\ref{#1}}

\def\1{\bm{1}}

\DeclareMathAlphabet{\mathsfit}{\encodingdefault}{\sfdefault}{m}{sl}
\SetMathAlphabet{\mathsfit}{bold}{\encodingdefault}{\sfdefault}{bx}{n}

\usepackage{hyperref}
\usepackage{url}

\usepackage{subcaption}
\usepackage{graphicx}
\usepackage{paralist}

\title{Anomalous behaviour in loss-gradient based interpretability methods}

\author{Vinod Subramanian\\
Queen Mary University of London\\
\texttt{v.subramanian@qmul.ac.uk}\\
\And
Siddharth Gururani\\
Electronic Arts\\
\texttt{sgururani@ea.com} \\
\And
Emmanouil Benetos\\
Queen Mary University of London\\
\texttt{emmanouil.benetos@qmul.ac.uk}\\
\And
Mark Sandler\\
Queen Mary University of London\\
\texttt{mark.sandler@qmul.ac.uk}
}

\iclrfinalcopy 
\begin{document}

\maketitle

\begin{abstract}
 

Loss-gradients are used to interpret the decision making process of deep learning models. In this work, we evaluate loss-gradient based attribution methods by occluding parts of the input and comparing the performance of the occluded input to the original input. We observe that the occluded input has better performance than the original across the test dataset under certain conditions. Similar behaviour is observed in sound and image recognition tasks. We explore different loss-gradient attribution methods, occlusion levels and replacement values to explain the phenomenon of performance improvement under occlusion.
\end{abstract}

\section{Introduction}

Quantitative evaluation of interpretability methods usually involves the ranking of input features using an interpretability algorithm, occluding parts of the input based on the ranking and measuring the change in the output as a result. Different techniques can be found in \cite{samek_evaluating_2017,fong_interpretable_2017, petsiuk_rise_2018, kindermans_learning_2017, feng-etal-2018-pathologies, carter2021overinterpretation}. \cite{hooker_benchmark_2019} suggests that re-training a model from random initialization is required to have a more accurate evaluation of interpretability methods because the training and test data need to have similar distributions.


Most research focuses on the highest ranked input features and the effects of occluding them. If the interpretability algorithm is good at identifying important features for the deep learning model then occluding the highest ranked features should cause a larger decrease in performance compared to an inferior interpretability algorithm. \cite{hooker_benchmark_2019} show that removing the lowest ranked features for saliency map based attribution methods causes the perform to degrade slower. \cite{kim_bridging_2019} perform experiments with the lowest ranked features of loss-gradient based methods but they do not report any results about performance improvements. \cite{ancona_towards_2018} show an example where occluding the lowest ranked features increases the pre-softmax activation, however they do not investigate it further.

Our work evaluates loss-gradient based attribution methods, focusing on how different occlusion levels and replacement values impact the test accuracy for a given model and attribution method. We focus on the highest and lowest ranked features. Our code can be found at \footnote{https://github.com/VinodS7/investigate-gradients} and the results can be summarised as follows:
\begin{compactenum}
    \item Removing the lowest ranked inputs can cause the performance to improve over the unchanged input;
    \item The sign of the gradients is important to the ranking process which seems counter-intuitive to the idea that the sign of the gradient just indicates the direction, not the importance of the input; 
    \item Replacing the input with different values changes the performance and appears related to the sign of the gradients.
\end{compactenum}


\section{Loss-gradient attribution methods}
\cite{kim_bridging_2019} show that loss-gradients are perpendicular to the decision boundary. Additionally, in adversarial attack literature the input is perturbed in the direction of the loss gradient in order to change the prediction \citep{szegedy_intriguing_2013, goodfellow_explaining_2014}. Given that our eventual goal is to bridge the gap between adversarial robustness and interpretability we decided to use loss-gradient based methods over saliency based methods.

The loss-gradient is the gradient of the loss with respect to the input to the model. For a model $D$ which has input $x\in\mathbb{R}^{m\times n}$ with ground truth label $t$ and loss function $L$, the loss-gradient $g\in\mathbb{R}^{m\times n}$ can be computed as:
\begin{equation}\label{eq:loss-grad}
    g = \frac{dL(D(x), t)}{dx}
\end{equation}
Based on the loss-gradient computation we use three attribution methods and we compare them to random removal of inputs. The first method is the unprocessed loss-gradients (grad\_orig) calculated in equation \ref{eq:loss-grad}. The second method is the absolute value of the gradients (abs\_grad) motivated by \cite{hooker_benchmark_2019} who observed slight performance improvements with the absolute value of the gradients over the raw values. The third method is multiplying the gradient with the input (grad\_inp) motivated by \cite{shrikumar_learning_2019}. The equations for the last two methods are:
\begin{align*}
    \text{Absolute gradient (abs\_grad)} &= \left|\frac{dL(D(x), t)}{dx}\right| \\
    \text{Gradient $\times$ input (grad\_inp)} &= \frac{dL(D(x), t)}{dx} \times x
\end{align*}

Once the attributes are obtained they need to be ranked in order to determine what input features to occlude. We want the highest ranked attributes to correspond to the most important input features and the lowest ranked attributes to correspond to the least important input features. For the absolute gradient method the attributes are arranged in descending order so the highest valued attribute corresponds to the highest ranked attribute. The other two methods contain positive and negative valued attributes and depending on how the loss function is implemented the signs could be inverted. We use the loss functions implemented in pytorch \footnote{pytorch.org} and our experiments show that the negative gradients rank highest and the positive gradients rank lowest.

\section{Experimental Setup}

We conduct experiments on two tasks, singing voice detection \citep{humphrey_singing} as the audio recognition task and image recognition on MNIST dataset \citep{lecun2010mnist}. We use the model from \cite{schluter2015exploring} for singing voice detection, the output of this model has a single head and is passed through a sigmoid function. If the output is closer to 1 then the audio contains singing voice and if the output is closer to 0 then the audio does not contain singing voice. We use the Area Under the Receiver Operating Characteristic curve (AUROC) to evaluate the model as it bypasses the need to set a threshold for classification. We train 4 versions of this model and the average AUROC across the models is 0.961. The loss function for this model is the binary cross entropy.

We simplify MNIST to a binary class problem between the "0" and "1" digits to keep it comparable to the singing voice detection task and to simplify the analysis. So all digits except the "0" and "1" digits are removed from the training and testing dataset. We use the example models from pytorch \footnote{https://github.com/pytorch/examples/tree/master/mnist} for the experiments. The output of the model has two heads that are passed through logsoftmax. The performance is evaluated using test accuracy. We train 5 versions of this model and the average accuracy is 99.99\%. The loss function used is the negative log likelihood.

\section{Experiments and Results}

\subsection{Evaluating different attribution methods}

\begin{figure}[!htb]
     \centering
     \begin{subfigure}[b]{0.85\textwidth}
         \centering
         \includegraphics[width=\textwidth]{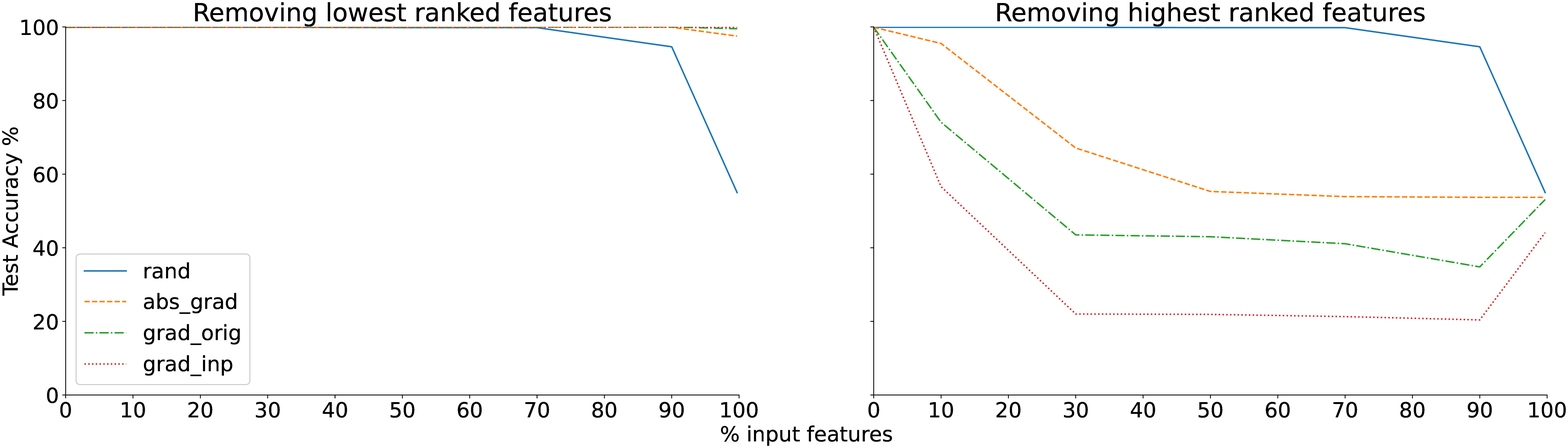}
         \caption{MNIST}
         \label{fig:mnist_att}
     \end{subfigure}
    \\
    \begin{subfigure}[b]{0.85\textwidth}
         \centering
         \includegraphics[width=\textwidth]{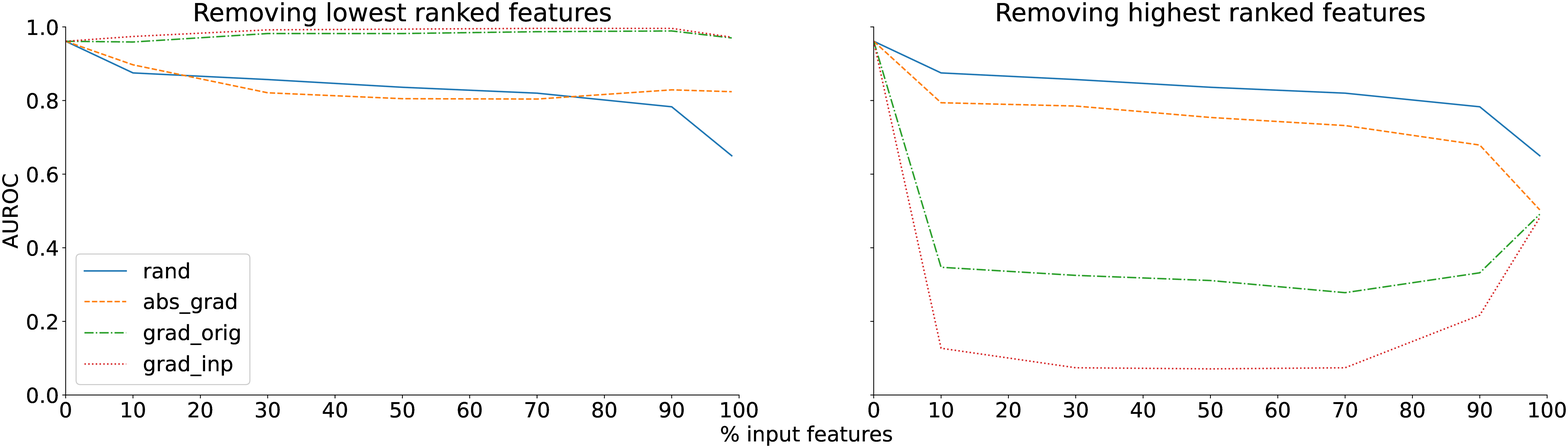}
         \caption{Singing Voice Detection}
         \label{fig:singing_att}
     \end{subfigure}
        \caption{Performance of attribution methods at different occlusion levels}
        \label{fig:attributions}
\end{figure}
We evaluate the model at 6 different occlusion levels, 10\%, 30\%, 50\%, 70\%, 90\% and a special case, for MNIST this is at 99.71\% which corresponds to 782 out of 784 pixels and for singing voice it is 98.91\% which corresponds to 9100 out of 9200 bins. The results are reported for the case of removing the highest ranked input features and the lowest ranked input features and the values are averaged across 5 models trained for each task. Figure \ref{fig:attributions} shows the results of this experiment. 

The behavior of the model after absolute gradient occlusion is as expected. Removing the highest ranked gradients causes the performance to degrade. Removing the lowest ranked features causes negligible performance change in the MNIST dataset until 99.71\% pixels are occluded, for singing voice detection the performance appears to decrease and then increase while overall performing much worse than without occlusion. The grad\_orig method performs better than the abs\_grad, where we observe that in both MNIST and singing voice detection the performance improves over normal evaluation for certain occlusion levels when removing the least important gradients. Finally, the grad\_inp degrades the performance more than the grad\_orig while removing the highest ranked features. For grad\_orig and grad\_inp the performance is worse than random chance while removing the highest ranked features which suggests that the occluded input consistently fools the classifier into predicting the wrong label, the increase in AUROC for singing voice detection indicates that the model is moving closer to a random classifier because informative features are occluded.

\begin{figure}[!htb]
     \centering
     \begin{subfigure}[b]{0.09\textwidth}
         \centering
         \includegraphics[width=\textwidth]{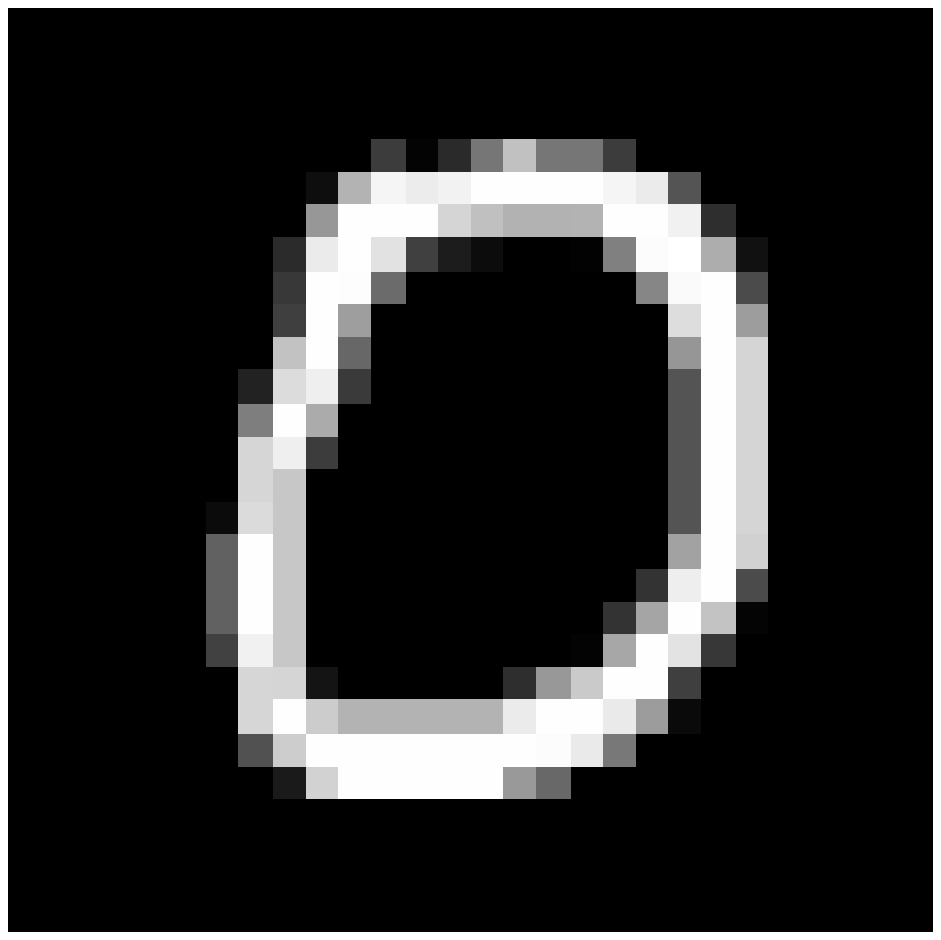}
     \end{subfigure}
     \hfill
     \begin{subfigure}[b]{0.09\textwidth}
         \centering
         \includegraphics[ width=\textwidth]{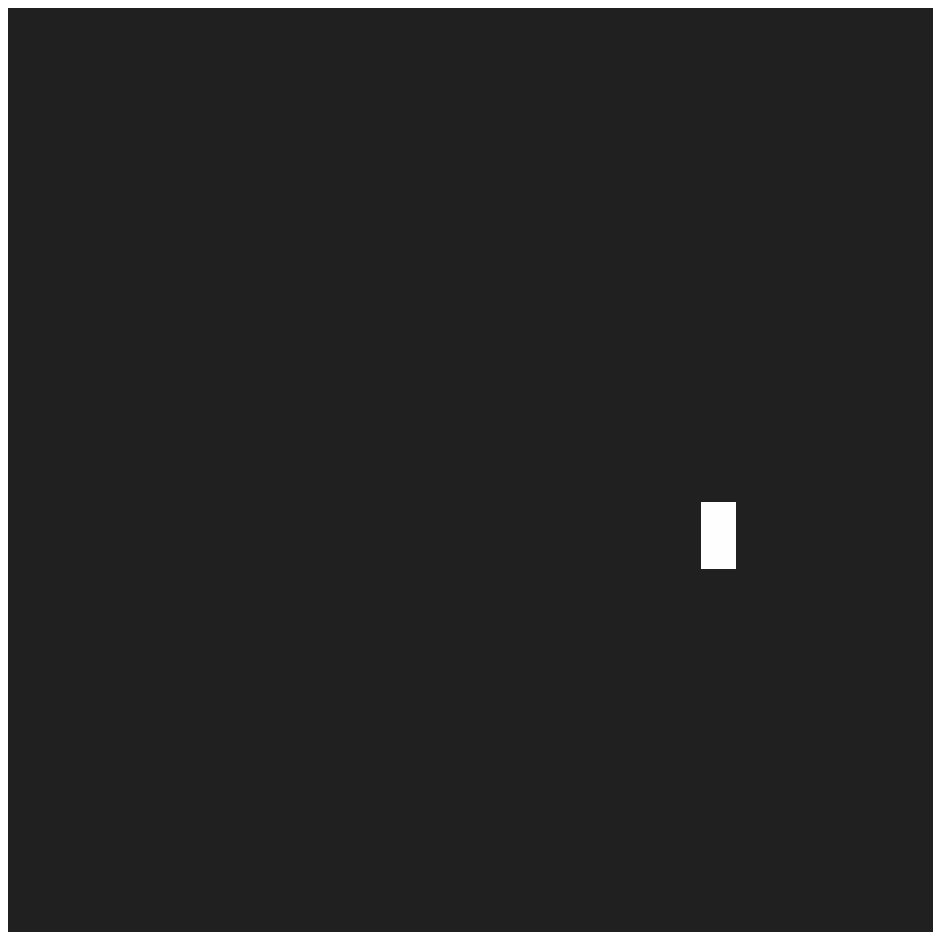}
     \end{subfigure}
     \hfill
     \begin{subfigure}[b]{0.09\textwidth}
         \centering
         \includegraphics[width=\textwidth]{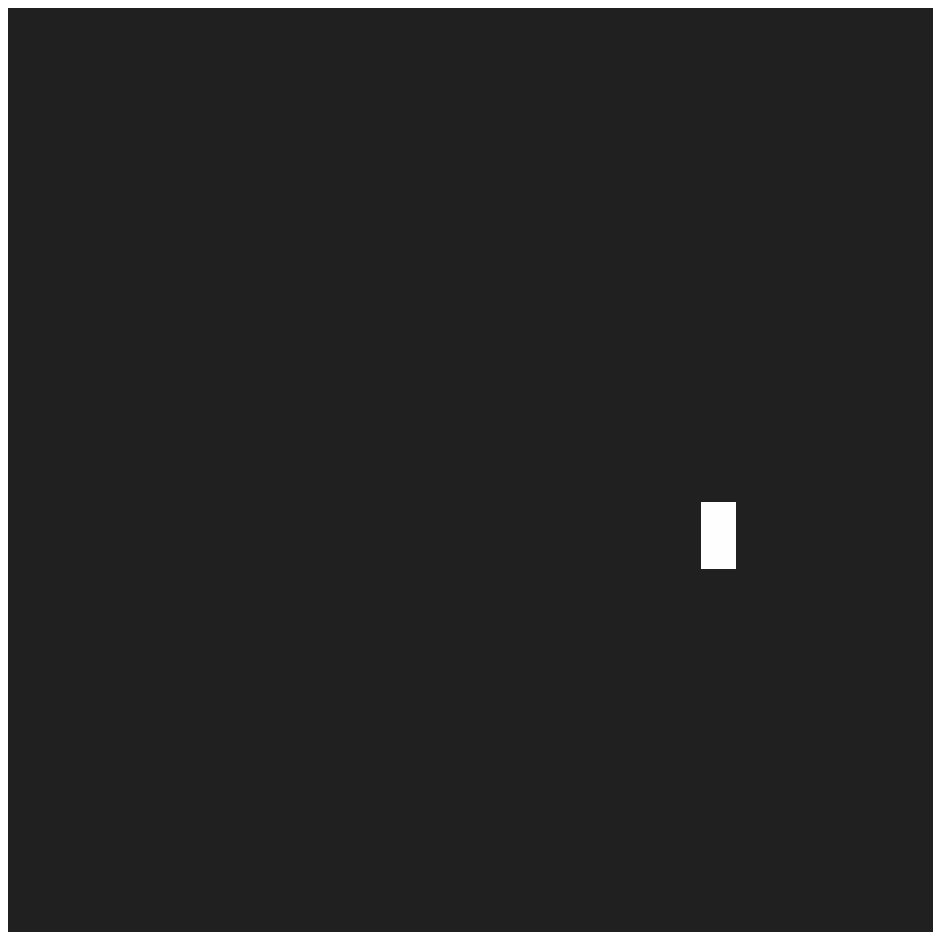}
     \end{subfigure}
     \hfill
     \begin{subfigure}[b]{0.09\textwidth}
         \centering
         \includegraphics[width=\textwidth]{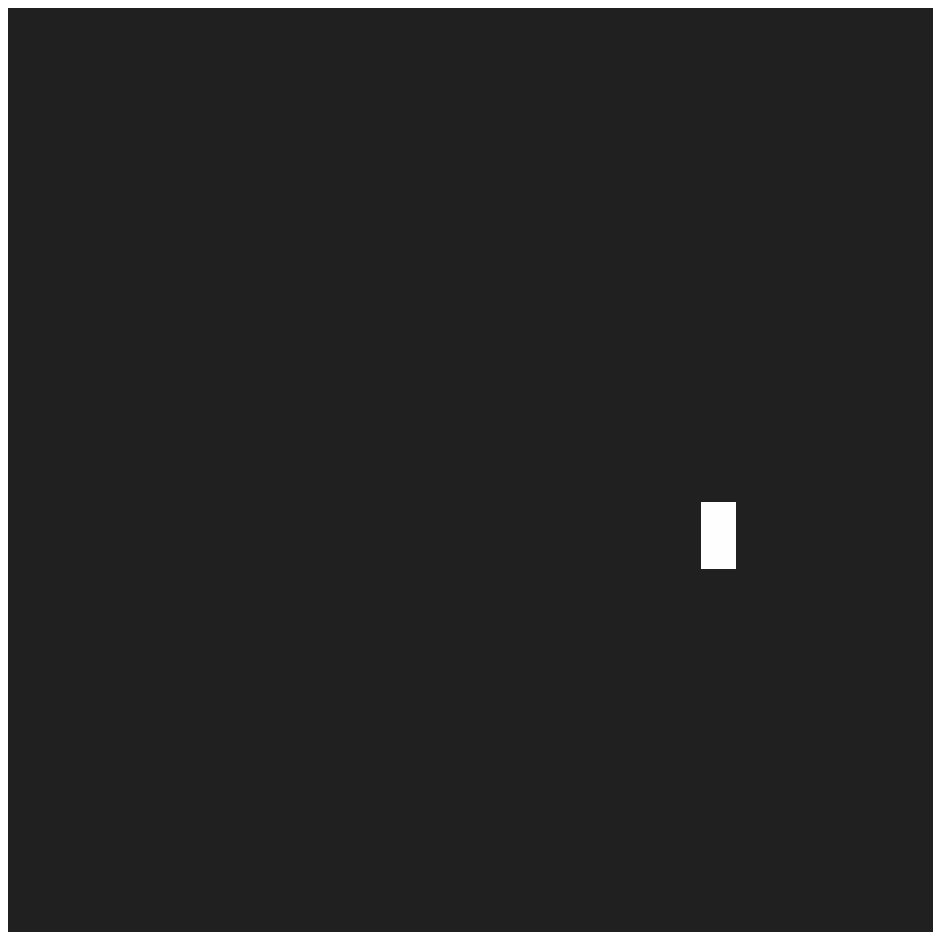}
     \end{subfigure}
    \hfill
    \begin{subfigure}[b]{0.09\textwidth}
         \centering
         \includegraphics[width=\textwidth]{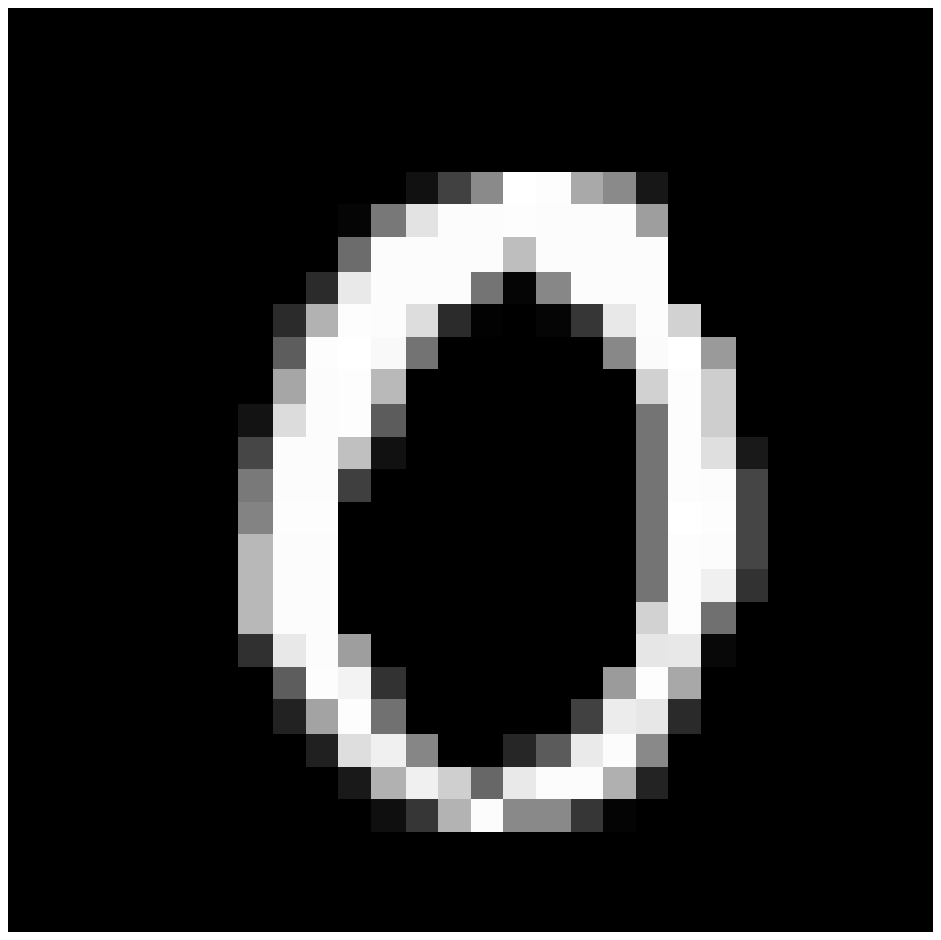}
     \end{subfigure}
     \hfill
     \begin{subfigure}[b]{0.09\textwidth}
         \centering
         \includegraphics[ width=\textwidth]{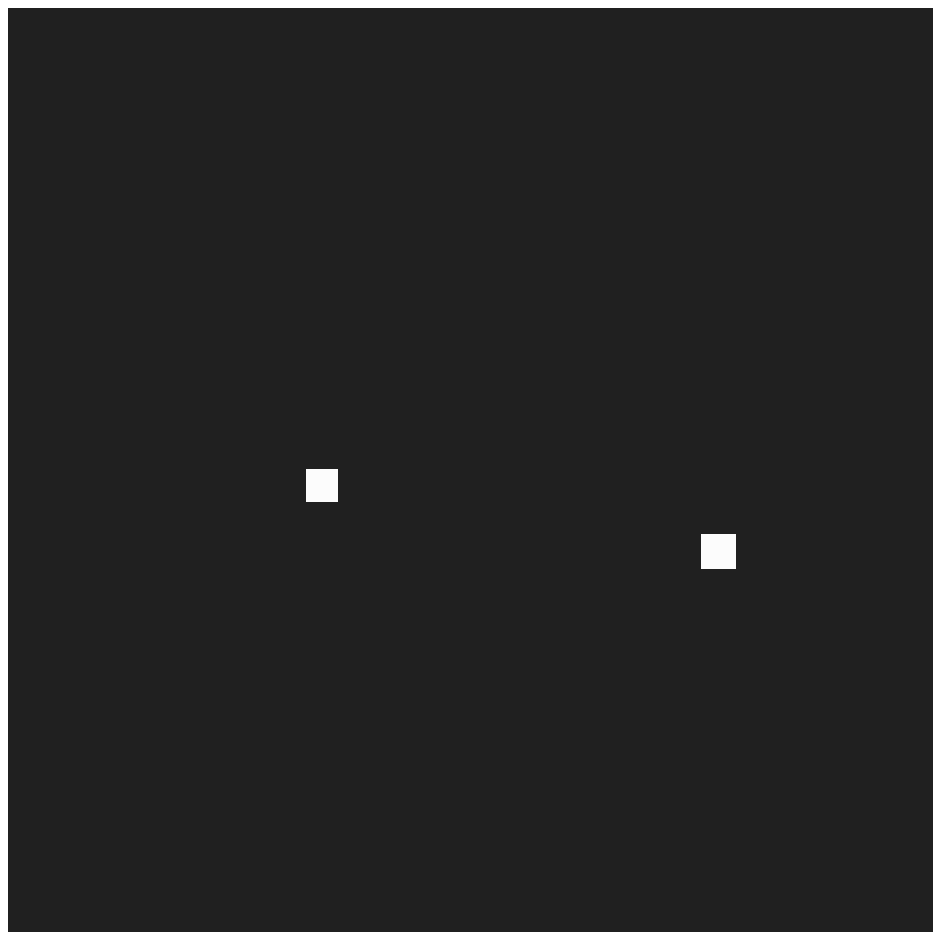}
     \end{subfigure}
     \hfill
     \begin{subfigure}[b]{0.09\textwidth}
         \centering
         \includegraphics[width=\textwidth]{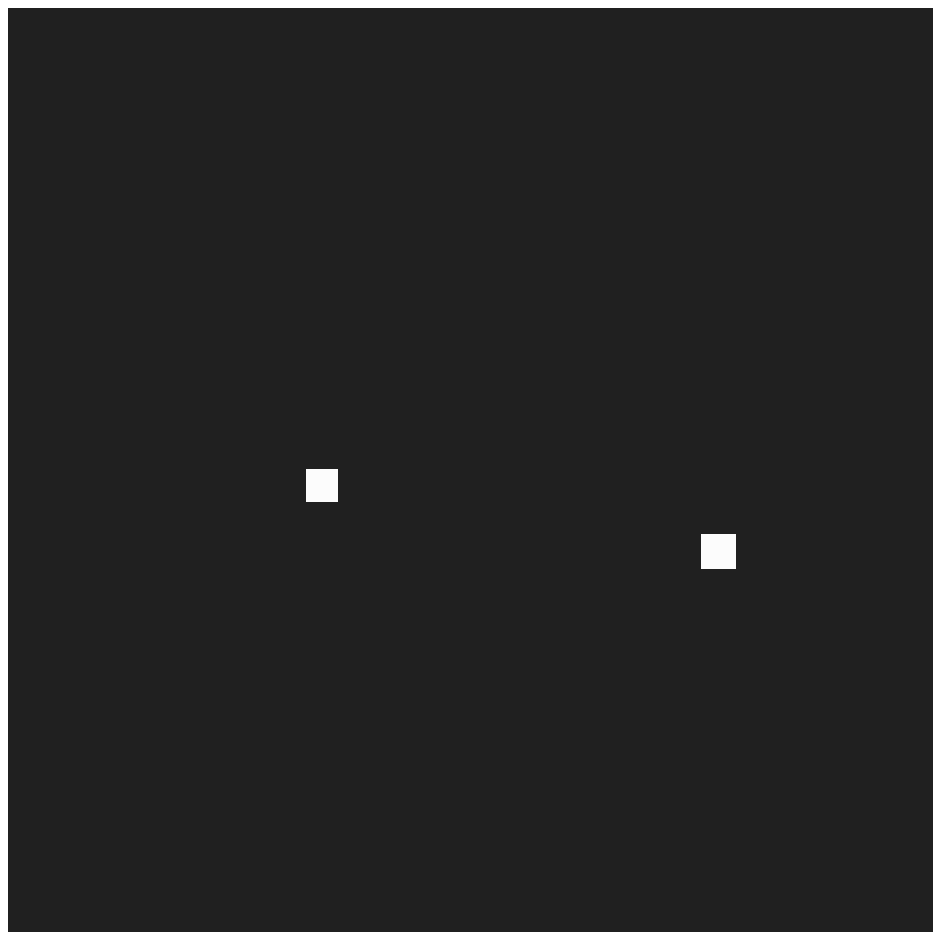}
     \end{subfigure}
     \hfill
     \begin{subfigure}[b]{0.09\textwidth}
         \centering
         \includegraphics[width=\textwidth]{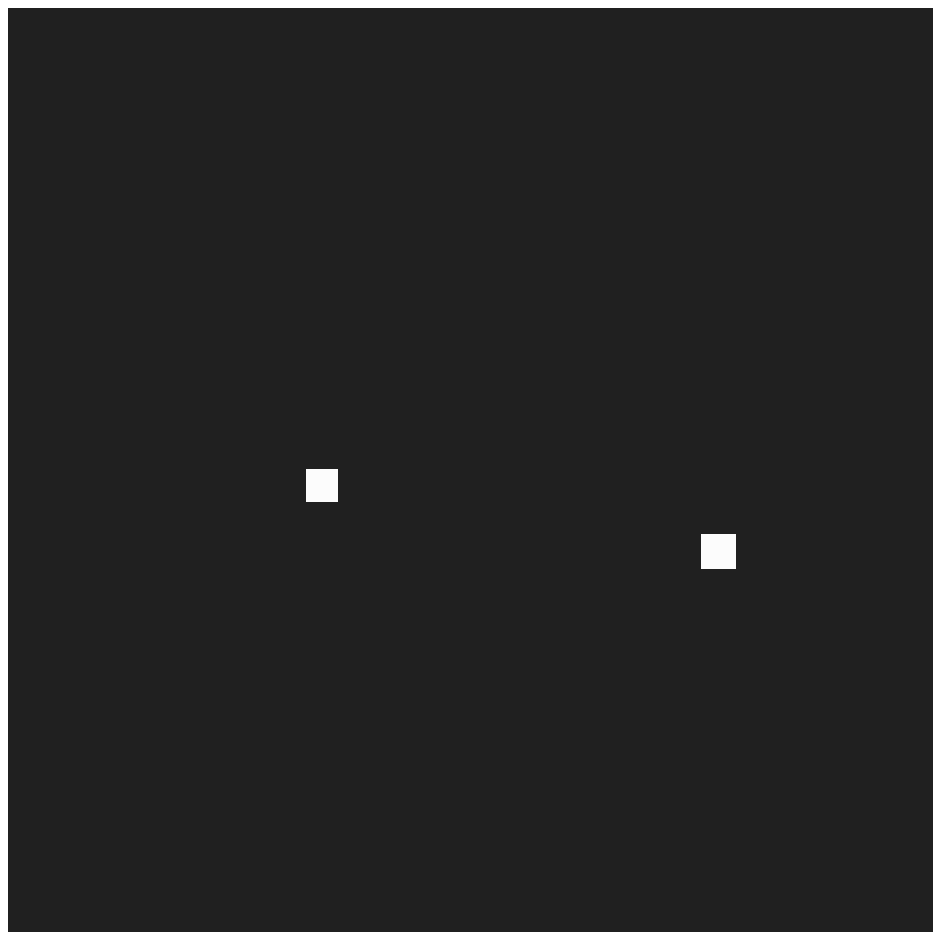}
     \end{subfigure}
     
     \begin{subfigure}[b]{0.09\textwidth}
         \centering
         \includegraphics[width=\textwidth]{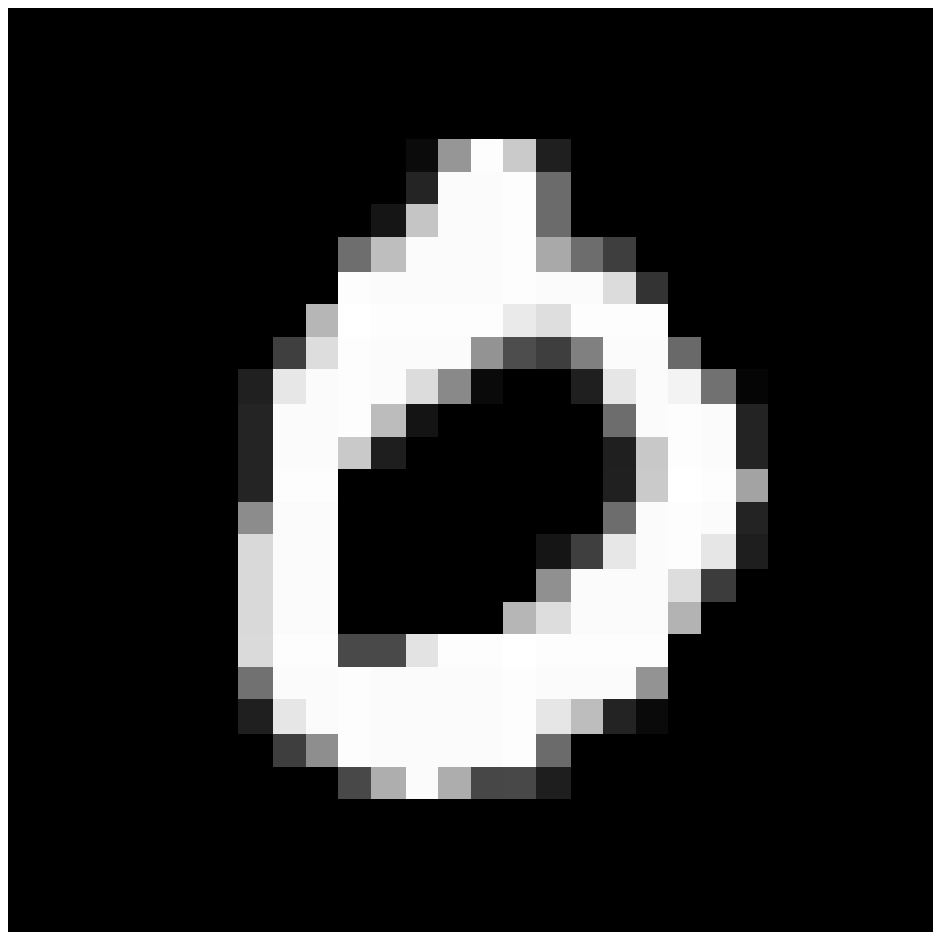}
         \caption*{original}
     \end{subfigure}
     \hfill
     \begin{subfigure}[b]{0.09\textwidth}
         \centering
         \includegraphics[ width=\textwidth]{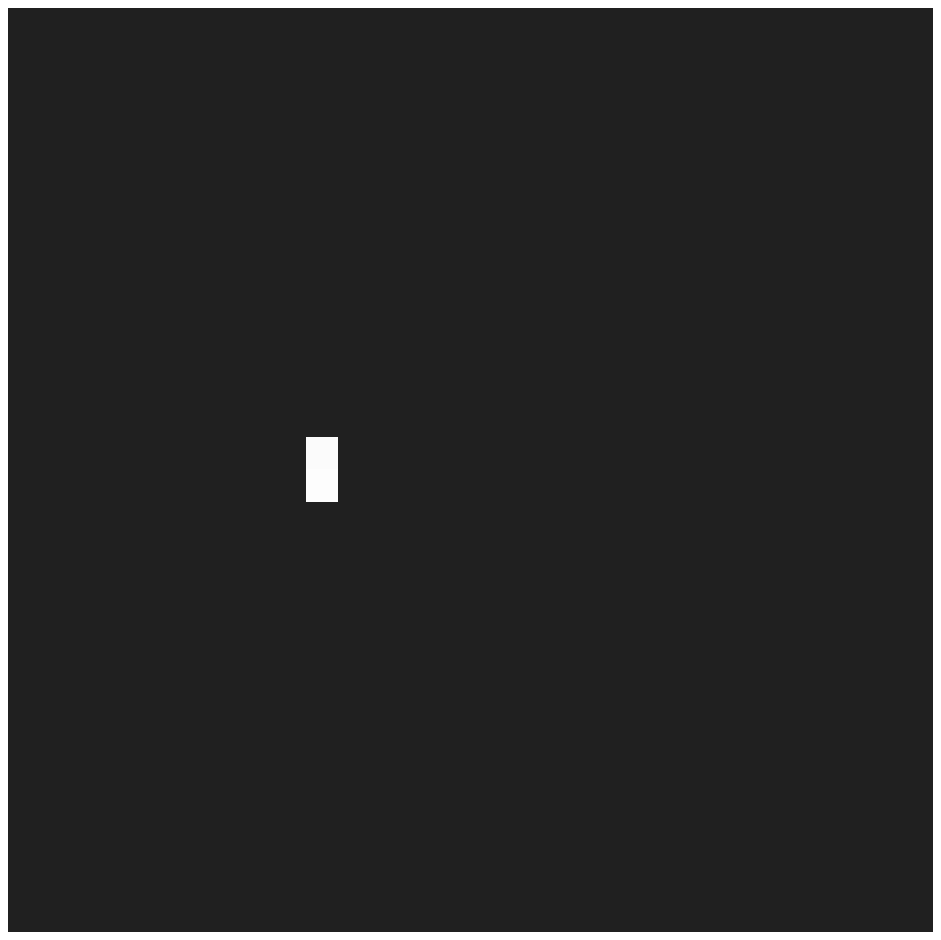}
         \caption*{abs\_grad}
     \end{subfigure}
     \hfill
     \begin{subfigure}[b]{0.09\textwidth}
         \centering
         \includegraphics[width=\textwidth]{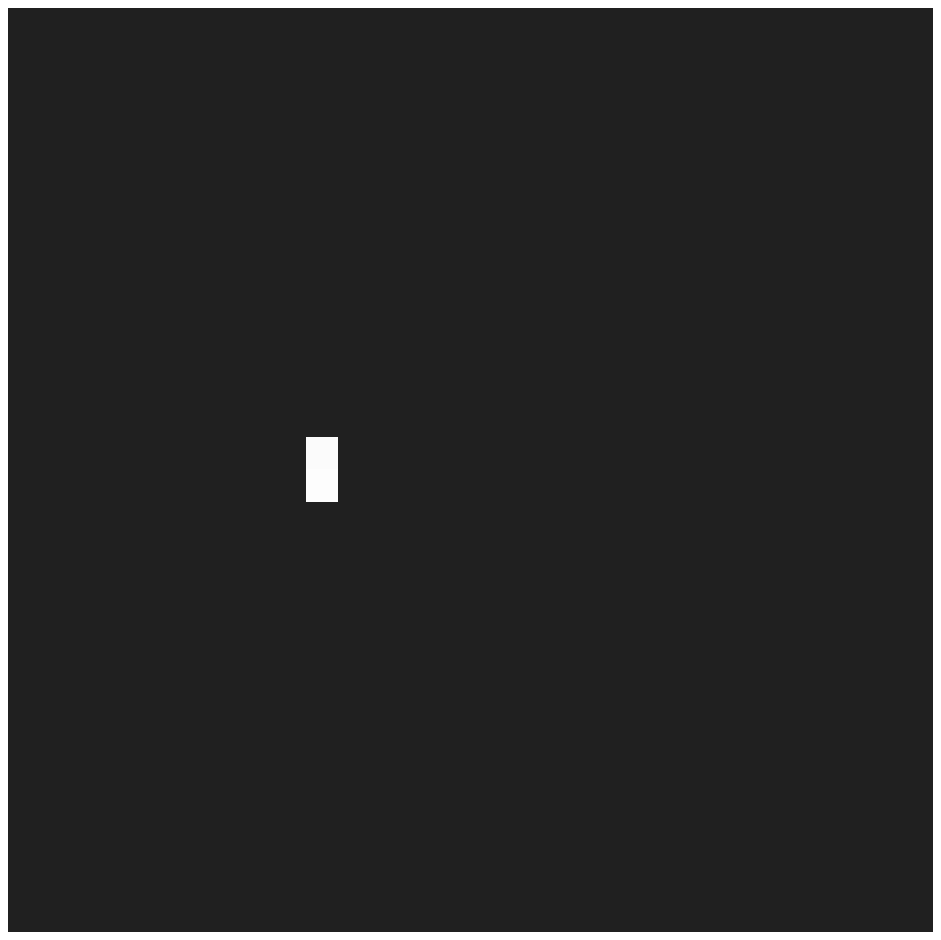}
         \caption*{grad\_orig}
     \end{subfigure}
     \hfill
     \begin{subfigure}[b]{0.09\textwidth}
         \centering
         \includegraphics[width=\textwidth]{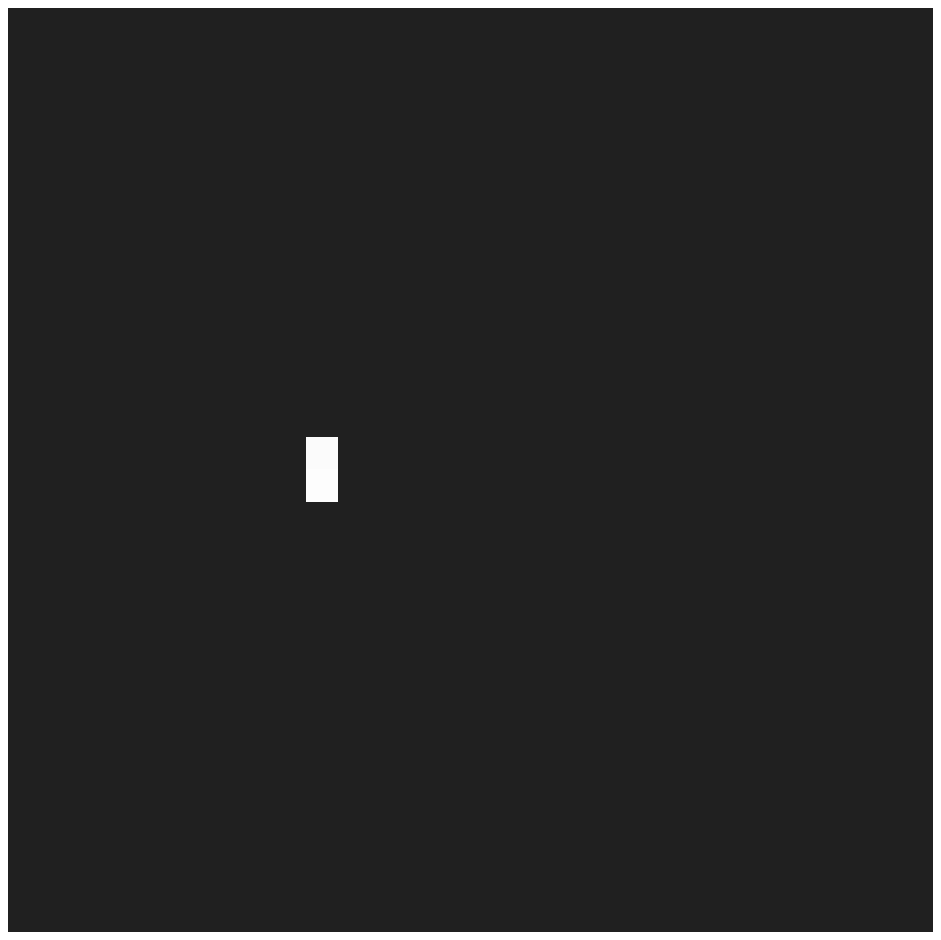}
         \caption*{grad\_inp}
     \end{subfigure}
    \hfill
    \begin{subfigure}[b]{0.09\textwidth}
         \centering
         \includegraphics[width=\textwidth]{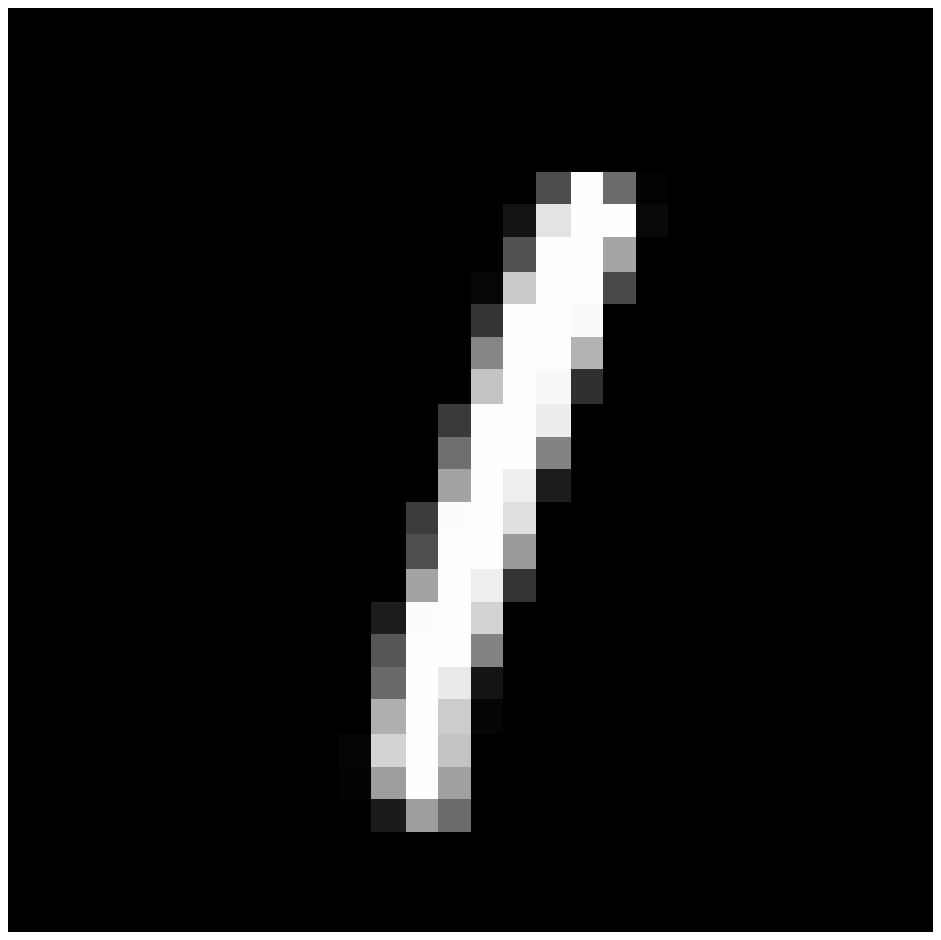}
         \caption*{original}
     \end{subfigure}
     \hfill
     \begin{subfigure}[b]{0.09\textwidth}
         \centering
         \includegraphics[ width=\textwidth]{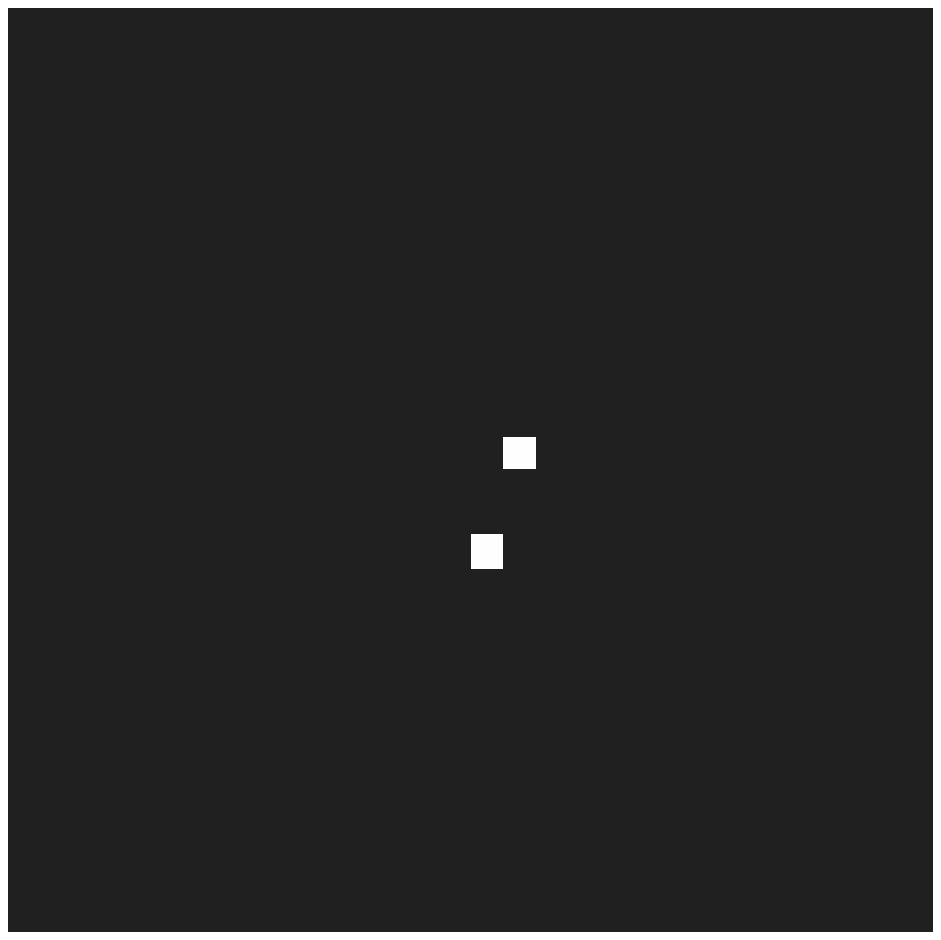}
         \caption*{abs\_grad}
     \end{subfigure}
     \hfill
     \begin{subfigure}[b]{0.09\textwidth}
         \centering
         \includegraphics[width=\textwidth]{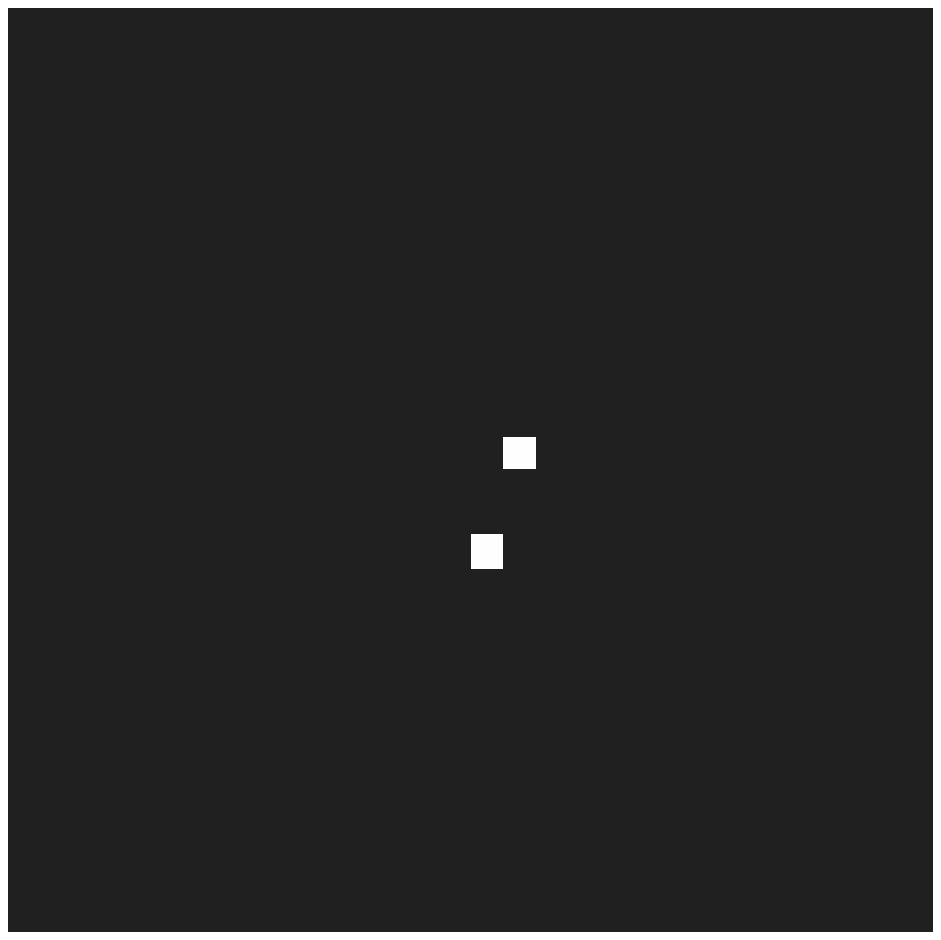}
         \caption*{grad\_orig}
     \end{subfigure}
     \hfill
     \begin{subfigure}[b]{0.09\textwidth}
         \centering
         \includegraphics[width=\textwidth]{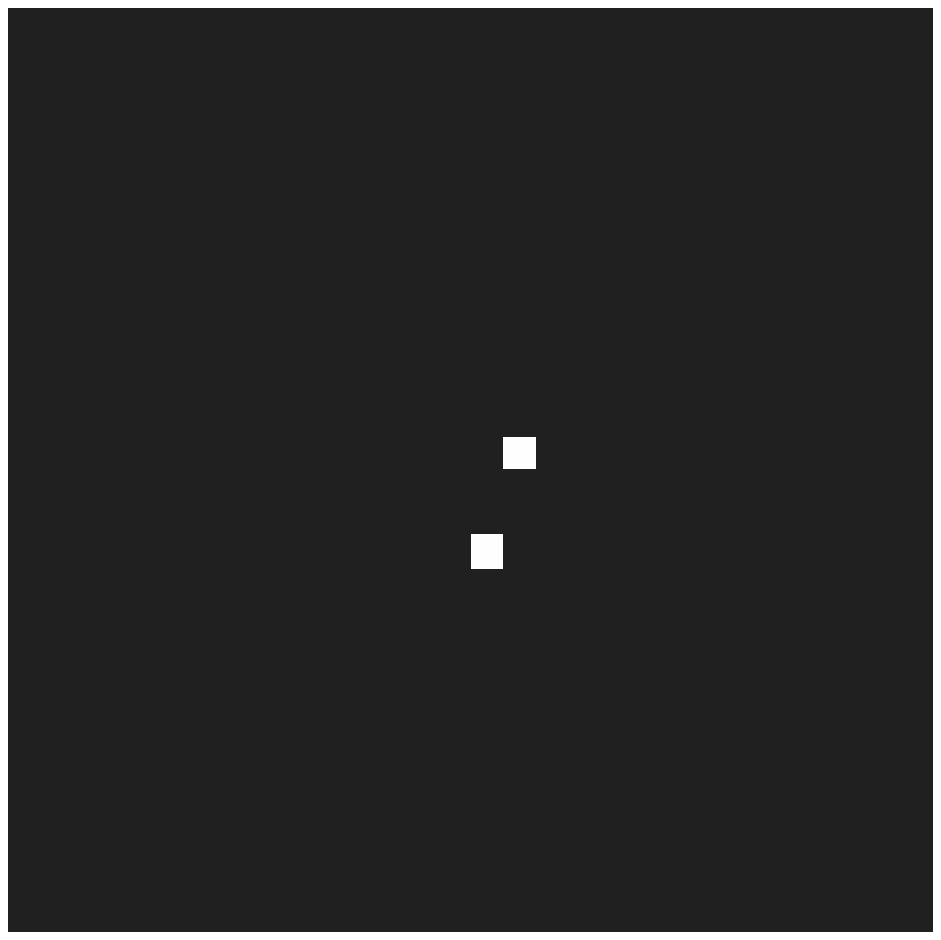}
         \caption*{grad\_inp}
     \end{subfigure}
        \caption{Comparison when 782 highest ranked pixel values are occluded by the dataset mean.}
        \label{fig:three graphs}
\end{figure}

We visualize the scenario where 99.71\% of the lowest ranked features of the input is occluded on the MNIST dataset in Figure \ref{fig:three graphs}. We observe a clear pattern where the central white pixels remain for the number 1 and the off centre white pixels remain for the number 0. Irrespective of the attribution method the two pixels remaining are the same. These examples suggest that there is an over emphasis on a small cluster of pixels to make a prediction. 

\subsection{Evaluating different replacement values}

\begin{figure}[!htb]
     \centering
         \includegraphics[width=0.85\textwidth]{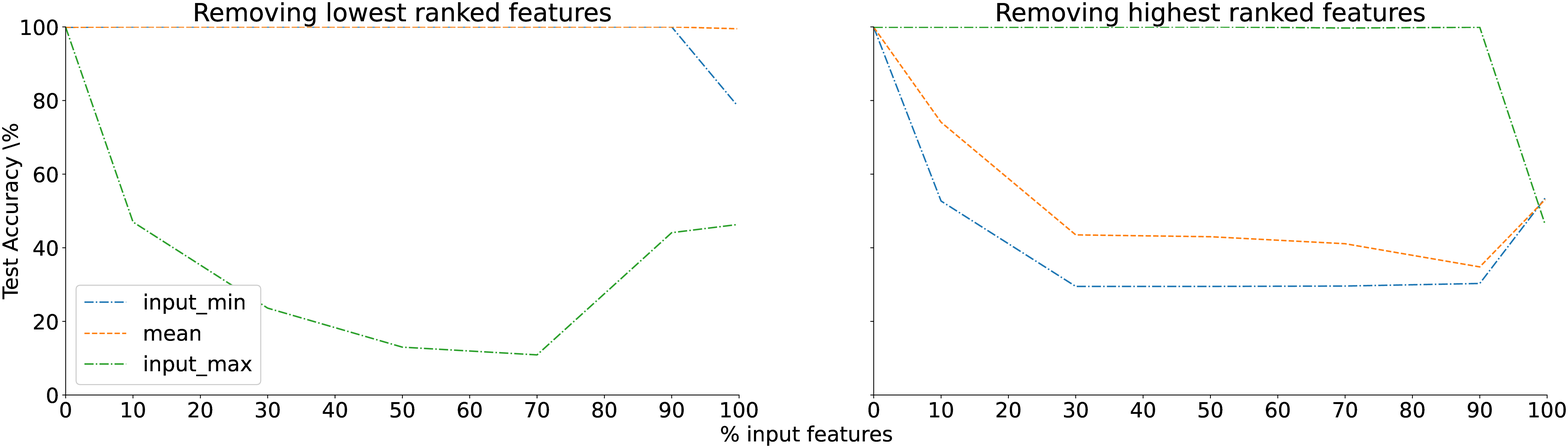}
         \caption{MNIST}
        \caption{Performance of different replacement values at different occlusion levels}
        \label{fig:replace_summary}
\end{figure}

In the previous experiment the occluded features in the input were replaced by the average value of the dataset. In this experiment we change the values that we are occluding the input with. We evaluate the performance at different occlusion levels for the grad\_orig method by replacing the features with the input minimum, input maximum and dataset mean. The results are shown in figure \ref{fig:replace_summary}. Replacing the occluded parts with the input minimum and mean has roughly similar behaviour for both tasks. Replacing by the input maximum inverts the behaviour of the model towards occluding by the highest and lowest ranked features.

\begin{figure}[!htb]
     \centering
     \begin{subfigure}[b]{0.15\textwidth}
         \centering
         \includegraphics[width=\textwidth]{images/orig_img_2.eps}
         \caption{original}
         \label{fig:orig}
     \end{subfigure}
     \hfill
     \begin{subfigure}[b]{0.15\textwidth}
         \centering
         \includegraphics[ width=\textwidth]{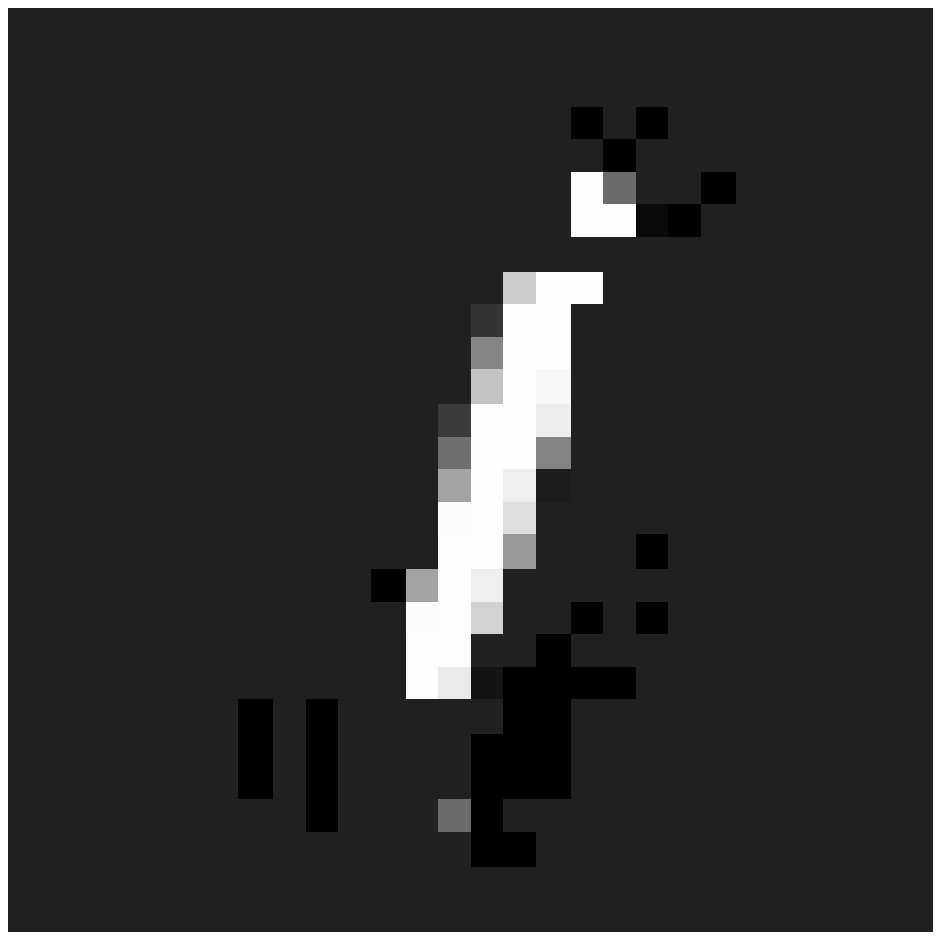}
         \caption{low mean}
         \label{fig:lowmean}
     \end{subfigure}
     \hfill
     \begin{subfigure}[b]{0.15\textwidth}
         \centering
         \frame{\includegraphics[width=\textwidth]{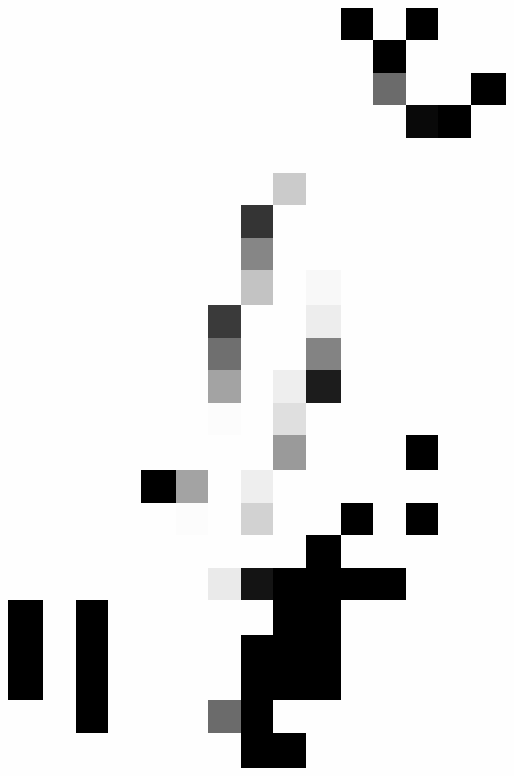}}
         \caption{low max}
         \label{fig:lowmax}
     \end{subfigure}
     \hfill
     \begin{subfigure}[b]{0.15\textwidth}
         \centering
         \includegraphics[width=\textwidth]{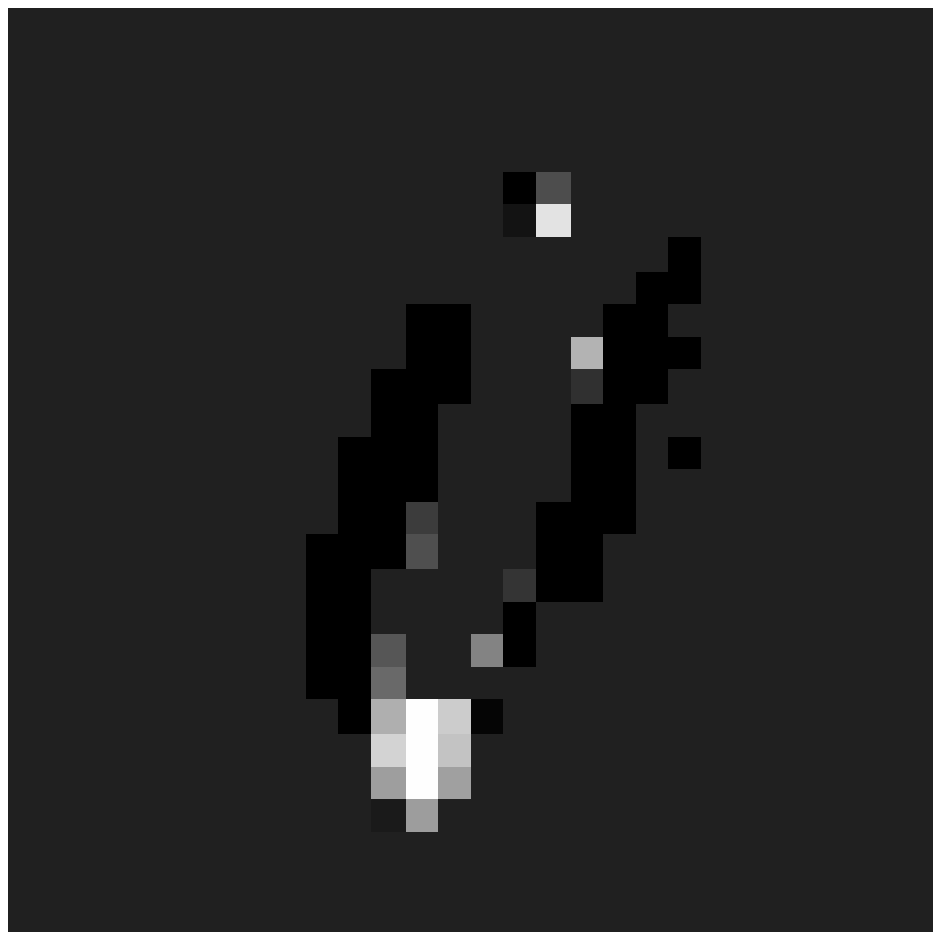}
         \caption{high mean}
         \label{fig:highmean}
     \end{subfigure}
     \hfill
     \begin{subfigure}[b]{0.15\textwidth}
         \centering
         \frame{\includegraphics[width=\textwidth]{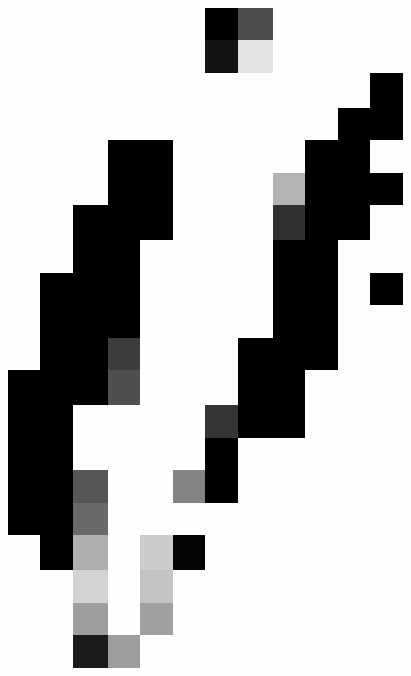}}
         \caption{high max}
         \label{fig:highmax}
     \end{subfigure}
        \caption{Replacing with mean and input maximum (max) of 90\% of lowest ranked (low) and highest ranked (high) features}
        \label{fig:replace}
\end{figure}

In the two pixel scenario for MNIST we observe that replacing by the input minimum instead of the mean the accuracy drops from 99.5\% to 78.7\% which suggests that the replacement values are important. In figure \ref{fig:replace} we compare the behaviour of replacing the highest ranked and lowest ranked features with the input maximum and the mean. In image \ref{fig:lowmean} the lowest ranked features are replaced by the mean the output is predicted correctly as "1", we could infer that those central white pixels are important for prediction. Replacing by the input maximum in image \ref{fig:lowmax} changes the prediction to "0" even though the same central white pixels are present. In image \ref{fig:highmean} we replace the highest ranked pixels with the mean and the prediction is "0" however replacing with the input maximum in image \ref{fig:highmax} is correctly predicted as "1". The similarity between image \ref{fig:lowmean} and \ref{fig:highmax} is that the edge between the white and dark pixels is preserved.

\section{Conclusion}
In this work we evaluate loss-gradient based attribution methods. We observe that the sign of the attribution method is important in the ranking process and the replacement value determines how the removed input features affect model placement. Our results suggest that the most important factor is the preservation of edges of the object. Future work will continue to explore the relationship of the replacement values with the occluded input features and also try to explain why the gradient $\times$ input method outperforms the raw gradients as an attribution method.
\section*{Acknowledgements}
This work has received funding from the EU’s Horizon 2020 research and innovation programme under the Marie Skłodowska-Curie grant agreement No. 765068.

\bibliography{robustml,adversarial-attacks}
\bibliographystyle{iclr2021_conference}

\newpage
\appendix
\section{Appendix}

\subsection{MNIST sigmoid model and normal MNIST}
To bridge the gap between the logsoftmax MNIST model and the sigmoid singing voice detection model we create a sigmoid MNIST model with a single output where the output 1 is assigned to the label "1" and the output 0 is assigned to the label "0". We set a threshold of 0.5 for classification. We also redo the experiments for the 10-class MNIST dataset to show that this is not a phenomenon that is unique to binary classification problem.The average accuracy of the 10 class logsoftmax model is 99.03\%, of the 2 class logsoftmax model is 99.95\% and of the sigmoid model is 99.95\%. The results are shown in table \ref{tab:occ_extra}, in all these results the occluded inputs are replaced with the dataset mean.

\begin{table}[!htb]
\caption{Performance for different occlusion levels. For the last 3 columns, the value on the left corresponds to the accuracy when removing the lowest ranked features and the value on the right is the accuracy when removing highest ranked features}
    \begin{subtable}{\linewidth}
      \centering
        \caption{MNIST LogSoftmax model }
        \begin{tabular}{|l|l|l|l|l|}
        \hline
        Occlusion \% & random & abs\_grad & grad\_orig & grad\_inp  \\ \hline
        10 & 0.999& 0.999 / 0.955& 1.0  / 0.741& 1.0 / 0.566  \\ \hline
        30 & 0.999& 0.999 / 0.671& 1.0 / 0.435& 1.0 / 0.220\\ \hline
        50 & 0.998& 0.999 / 0.553& 1.0 / 0.430& 1.0 / 0.219\\ \hline
        70 & 0.998& 0.999 / 0.539& 1.0 / 0.411& 1.0 / 0.213\\ \hline
        90 & 0.946& 0.999 / 0.537 & 1.0 / 0.348 & 1.0 / 0.204\\ \hline
        99.71 & 0.550& 0.975 / 0.537& 0.995 / 0.531& 0.999 / 0.441 \\ \hline
        \end{tabular}
    \end{subtable}%
    \hfill
    \begin{subtable}{\linewidth}
      \centering
        \caption{MNIST sigmoid model}
        \begin{tabular}{|l|l|l|l|l|}
        \hline
        Occlusion \% & random& abs\_grad & grad\_orig & grad\_inp \\ \hline
        10 & 0.999&  0.999 / 0.959 & 1.0 / 0.710 & 1.0 / 0.567 \\ \hline
        30 & 0.999& 0.999 / 0.665  & 1.0 / 0.393  & 1.0 / 0.276 \\ \hline
        50 & 0.999& 0.999 / 0.556 & 1.0 / 0.390 & 1.0 / 0.274\\ \hline
        70 & 0.998& 0.999 / 0.539 & 1.0 / 0.384 & 1.0 / 0.275 \\ \hline
        90 & 0.970& 0.999 / 0.537 & 0.999 / 0.350 & 0.999 / 0.277  \\ \hline
        99.71 & 0.571& 0.982 / 0.537 & 0.999 / 0.517 & 1.0 / 0.397 \\ \hline
        \end{tabular}
    \end{subtable}
    \hfill
    \begin{subtable}{\linewidth}
      \centering
        \caption{MNIST LogSoftmax 10}
        \begin{tabular}{|l|l|l|l|l|}
        \hline
        Occlusion \% & random& abs\_grad & grad\_orig & grad\_inp \\ \hline
        10 & 0.984&  0.990 / 0.693 & 0.996 / 0.585 & 0.995 / 0.567 \\ \hline
        30 & 0.928& 0.987 / 0.368  & 0.985 / 0.459  & 0.994 / 0.276 \\ \hline
        50 & 0.730& 0.979 / 0.252 & 0.978 / 0.416 & 0.994 / 0.274\\ \hline
        70 & 0.427& 0.952 / 0.188 & 0.965 / 0.362 & 0.993 / 0.275 \\ \hline
        90 & 0.205& 0.772 / 0.140 & 0.892 / 0.224 & 0.984 / 0.277  \\ \hline
        99.71 & 0.119& 0.150 / 0.115 & 0.220 / 0.114 & 0.265 / 0.397 \\ \hline
        \end{tabular}
    \end{subtable}
    \label{tab:occ_extra}
\end{table}

\end{document}